\documentclass[11pt]{article}

\usepackage[preprint]{acl}

\usepackage{times}
\usepackage{latexsym}

\usepackage[T1]{fontenc}

\usepackage[utf8]{inputenc}

\usepackage{microtype}

\usepackage{inconsolata}

\usepackage{graphicx}

\usepackage[normalem]{ulem} 

\usepackage{amsmath}
\usepackage{booktabs}

\title{Language Models Generalize to Human-like Word Order Preferences}

 \author{Amanda Popadich \and Shane Steinert-Threlkeld \\
         Department of Linguistics, University of Washington \\ \{ popadich, shanest \}@uw.edu }

\begin{document}

\maketitle

\begin{abstract}
A central question in language acquisition is whether linguistic biases can emerge from general learning mechanisms operating over underdetermined input. Artificial Language Learning (ALL) studies have shown that human learners reliably generalize beyond the evidence provided, including by preferring scope-homomorphic noun phrase modifier orders. In this work, we investigate whether language models exhibit the same bias under similar conditions. We create a controlled learning environment in which models are trained on a corpus where all noun phrases containing multiple modifiers have been removed, eliminating direct evidence about modifier ordering, and are then evaluated on multiple modifier sentences. Across three model sizes, we find that they consistently prefer scope-homomorphic orders despite never observing them during training. These preferences vary in strength by modifier type. To investigate the source of these preferences, we examine noun-modifier association strength using pointwise mutual information (PMI). While PMI reflects known modifier-ordering patterns, it does not explain the models' ordering preferences. These findings demonstrate that LMs can recover human-like linguistic generalizations from impoverished input and provide a controlled framework for investigating the mechanisms underlying such biases.
\end{abstract}

\section{Introduction}
A central question in linguistics is how learners acquire rich grammatical systems despite receiving limited and ambiguous input. Human learners routinely generalize beyond the evidence available to them, suggesting that language acquisition is guided by inductive biases that constrain how learners extrapolate from sparse data \citep{Pearl2022a,Crain1991}. A major goal of Artificial Language Learning (ALL) research is therefore to identify these biases by examining how learners generalize when multiple hypotheses are compatible with the observed input \citep{Culbertson2023}. A consistent finding from this literature is that learners prefer structure-based generalizations over those based solely on surface word order frequencies \citep{Culbertson.Adger2014,Culbertson.etal2020}.

Recent work has used neural language models (LMs) as model learners for investigating similar questions. This research shows that LMs acquire a wide range of syntactic generalizations and can generalize to phenomena for which direct evidence has been removed from training data \citep{Linzen.etal2016a,McCoy.etal2020a,Patil.etal2024,Warstadt.Bowman2020,Misra.Mahowald2024,Leong.Linzen2026}. Such experiments provide a controlled environment for testing whether linguistic biases require language-specific constraints or can emerge from experience.

\begin{figure*}[ht]
  \centering
  \includegraphics[width=\textwidth, height=4.75cm]{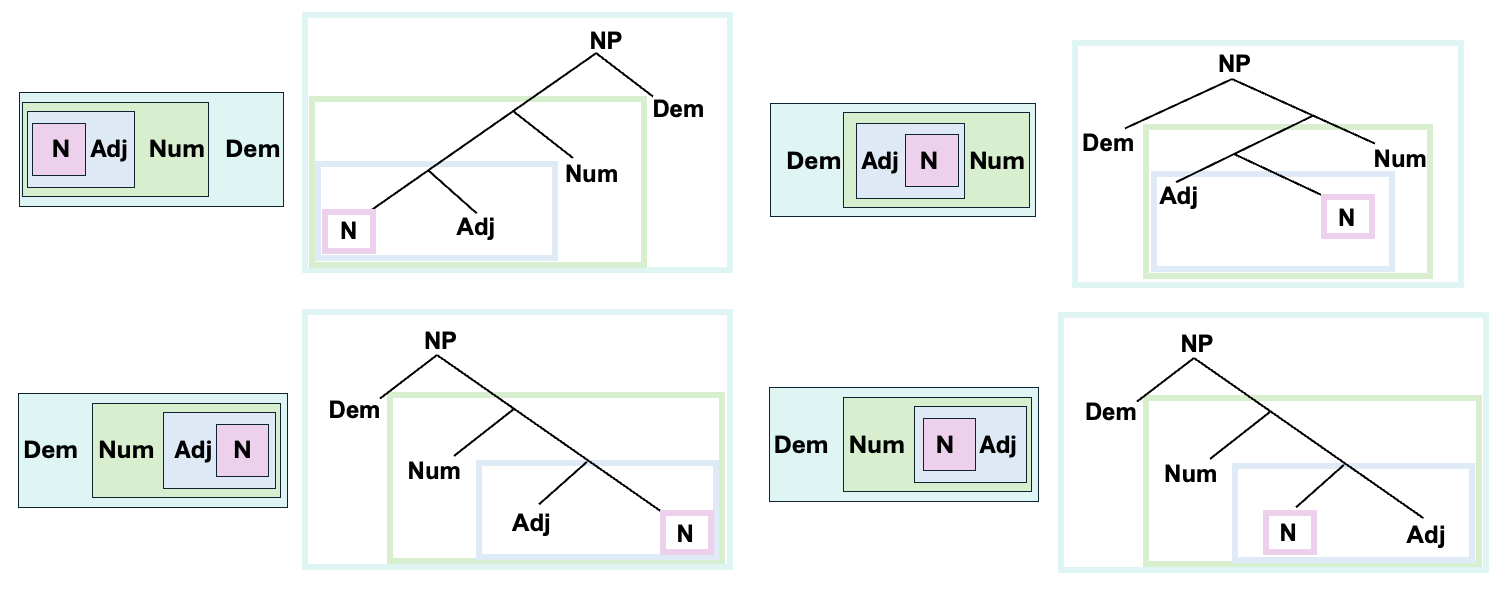}
  \caption{A few possible examples of scope-homomorphic noun phrase orderings. The trees on the right depict the hierarchical semantic structure that is realized as the linear word order shown on the left, with boxes indicating the semantic scope of each modifier.}
  \label{scope}
\end{figure*}

In this paper, we investigate one of the most well-established biases observed in ALL experiments: scope-homomorphism. This constraint predicts that modifiers within a noun phrase are linearly ordered in a manner that reflects their compositional semantic scope. For example, English permits \textit{two fluffy dogs} but not \textit{* fluffy two dogs}.

To determine whether language models exhibit this bias, we adapt the ALL paradigm using Filtered Corpus Training (FiCT; \citealp{Patil.etal2024}). We train models on a corpus where all multi-modifier noun phrases have been removed and replaced with single-modifier instances, eliminating direct evidence about modifier ordering, and test whether models nevertheless generalize to scope-homomorphic orders.

Our contributions are in three parts. First, we introduce a FiCT paradigm for investigating scope-homomorphic word order in language models that closely mirrors the poverty-of-the-stimulus conditions used in human ALL experiments (Section~\ref{sec:method}). Second, we show that language models consistently prefer scope-homomorphic orders despite never observing noun phrases with multiple modifiers during training across multiple model sizes (Section~\ref{sec:results}). Third, we compare these preferences with human experimental results and investigate possible distributional sources of the learned bias, finding important similarities as well as systematic differences in the strength and source of these preferences (Section~\ref{sec:bias-source}).  In particular, we show that pointwise mutual information does not predict model preferences.


\section{Background and Related Work}
\subsection{Artificial Language Learning}
A successful method for probing the cognitive biases that learners bring to language acquisition is the Artificial Language Learning paradigm \citep{Culbertson2023}. The Poverty of the Stimulus (or extrapolation) method exposes learners to a miniature language compatible with multiple rules, then tests their ability to generalize to unseen forms. The patterns of generalization produced provide insight into the inductive biases that guide language learning under conditions of underdetermined evidence. This paradigm has become a standard approach for investigating how humans extrapolate from limited input in poverty of the stimulus settings \citep[e.g.,][]{Wilson2006,Culbertson.Adger2014,Martin.etal2019}.

In this way, ALL studies aim to identify whether linguistic universals can emerge even in the absence of direct evidence, thereby offering insight into the nature of the constraints learners bring to the task. Within this framework, we focus on a specific proposed universal that is central to the present study.

\subsection{Scope-Homomorphism}
A key finding from ALL studies is that learners prefer structure-based generalizations over those based on surface order frequencies \citep{Culbertson2023}. Within word order, this pattern is particularly evident in preferences for \textit{scope-homomorphism}, where learners order modifiers according to their underlying semantic relationships rather than surface frequencies \citep{Greenberg1963,Culbertson.Adger2014,Martin.etal2019,Martin.etal2020}. Here, we look at the modifiers within the noun phrase. Figure \ref{scope} illustrates how these semantic relationships map onto linear order.

\citet{Culbertson.Adger2014} first investigated this word order universal using the ALL paradigm. Participants learned individual noun-modifier combinations but were never exposed to phrases containing multiple modifiers, leaving the relative ordering of modifiers underdetermined. When later asked to produce novel multi-modifier noun phrases, participants reliably preferred scope-homomorphic orders over alternative linear orders, with the strongest effects observed for demonstrative-adjective combinations.

Subsequent studies have shown that this preference persists despite potential concerns about transfer from participants’ native languages \citep{Martin.etal2020}. In particular, people continue to exhibit scope-homomorphic preferences even when their native language provides conflicting evidence \citep{Martin.etal2024} and when artificial languages are fully iconic \citep{Shapiro.Steinert-Threlkeld2023,Shapiro.etal2024, Tal.etal2025a}. These findings suggest that the preference reflects a robust inductive bias rather than simply learned properties of participants’ native languages.

However, while ALL experiments demonstrate that learners possess systematic biases in underdetermined learning contexts, they do not reveal the source of these biases. Such preferences could arise from innate linguistic constraints, prior language experience, or general learning mechanisms. Controlled learning environments are therefore needed to investigate which properties of the input are sufficient to produce these generalizations.

\begin{figure*}[ht]
  \centering
  \includegraphics[width=\textwidth, height=8cm]{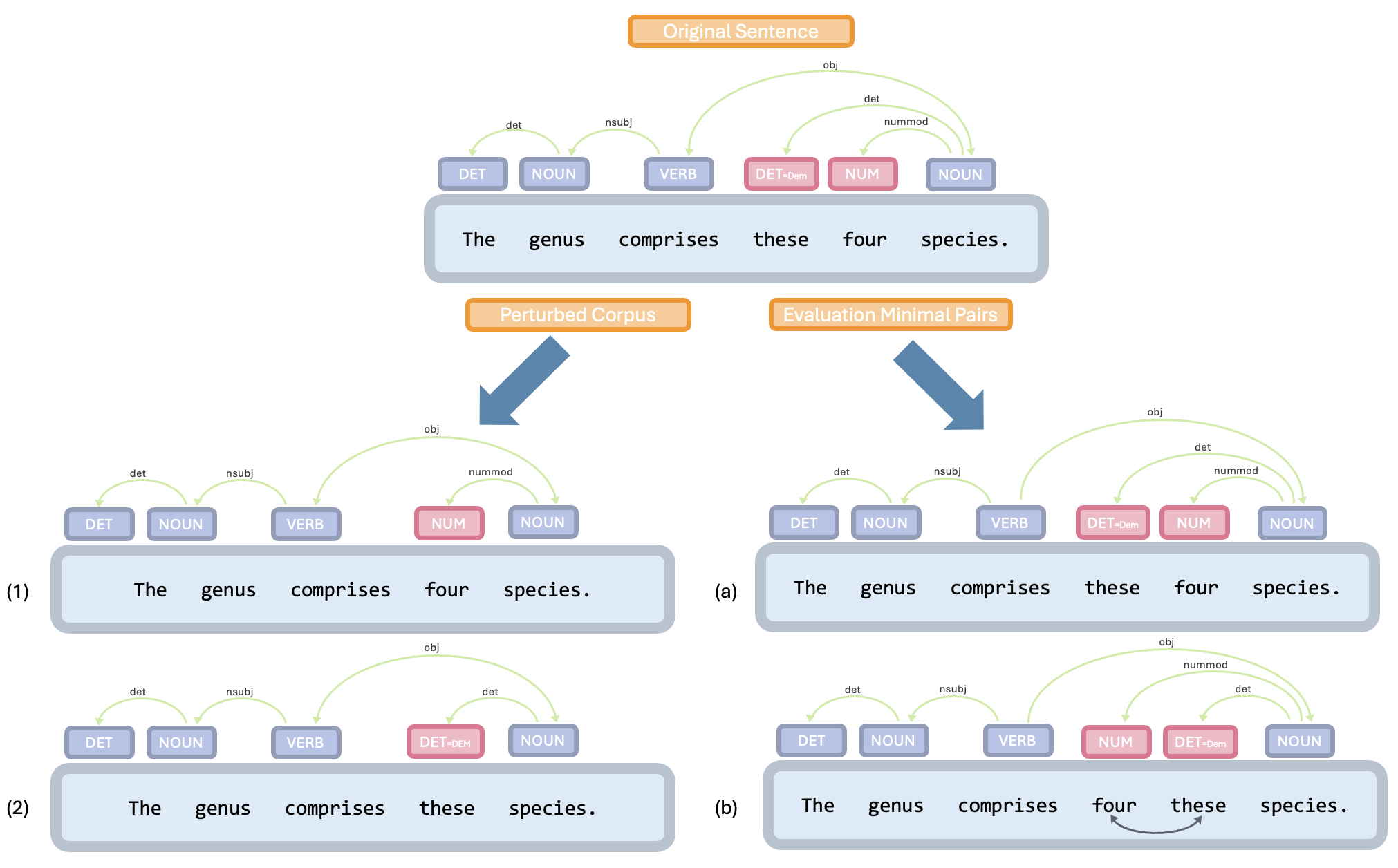}
  \caption{Example of the corpus perturbation procedure for a Dem-Num noun phrase. Multi-modifier noun phrases are decomposed into variants containing at most one modifier per noun. The minimal pair 1a and 1b compares the homomorphic and non-homomorphic modifier orders for the evaluation set.}
  \label{pipeline}
\end{figure*}

\subsection{Inductive Biases in Language Learning}
The Poverty of the Stimulus (PoS) argument holds that because linguistic input underdetermines the grammatical knowledge learners acquire, language learning must be guided by constraints that shape how learners generalize from experience \citep{Pearl2022a}. One hypothesis for how learners constrain their expectations comes from the nativist approach to linguistics, which posits that learners bring innate linguistic knowledge to language acquisition, thereby constraining the hypothesis space \citep{Chomsky1957, Chomsky1959, Crain.Nakayama1987}.

In contrast, constructivist and usage-based approaches argue that the richness of linguistic knowledge can be derived from general learning mechanisms that track distributional regularities in the input, without requiring extensive innate linguistic constraints \citep{Tomasello2003b,Bybee2006}. In this way, the domain general biases of the learner are sufficient to learn complex structure-based rules from indirect evidence \citep{Perfors.etal2011a,Reali.Christiansen2005}. To evaluate these claims, one needs an example of a domain-general learner who displays a preference for linguistic constraints despite having no language-specific innate bias.

\subsection{Language Models as Model Learners}
Recent advances with LMs have provided a new way to investigate the learning biases of general language learners. Prior work has shown that LMs of various sizes and architectures make structure-based syntactic generalizations \citep[e.g.][]{Linzen.etal2016a,Wilcox.etal2018,Gulordava.etal2018a,Goldberg2019,Hu.etal2020,Papadimitriou.etal2021,Wilson.etal2023,Warstadt.Bowman2020}, and have used models to probe other typological generalizations, including adjective ordering preferences \citep{Jumelet.etal2026}. Recent work has adopted controlled training conditions, such as FiCT, which removes specific phenomena from the input and evaluates whether models can recover the corresponding generalizations from indirect evidence. FiCT experiments have shown that LMs can generalize to unseen phenomena, including Negative Polarity Item Licensing \citep{Jumelet.etal2021a}, rare Article+Adjective+Numeral+Noun constructions \citep{Misra.Mahowald2024}, and a broad range of syntactic dependencies \citep{Patil.etal2024,Leong.Linzen2026}.

While FiCT provides a structured way of selectively removing evidence for targeted linguistic phenomena, it is also possible to alter or perturb training data to create input that follows a specific distribution to probe model generalization. Related work has used synthetic or perturbed corpora to study whether language models can acquire grammatical systems under altered input conditions, such as typologically atypical or highly constrained languages \citep{Kallini.etal2024b, Xu.etal2026}, as well as adding differential argument marking to English corpora \citep{Deng.etal2026}. 

While LMs are not intended to be direct models of human learners, they provide a useful framework for evaluating the sources of linguistic biases. In the following, we describe our experiment instantiating a top-down ALL experiment meant to probe whether the scope-homomorphism preference is learnable by these model learners. 

\section{Method}
\label{sec:method}

We investigate whether LMs learn scope-homomorphic modifier ordering preferences from impoverished input. We train models on a perturbed corpus without multi-modifier NPs and evaluate whether they recover these preferences using minimal-pair comparisons.

\subsection{Training Corpus}
We use the 2023 Wikipedia dump from the Wikimedia Foundation \citep{wikidump}, filtered to remove non-content sections and sentences with fewer than two tokens. We randomly sample sentences until reaching a 100M-token training corpus.

To build the perturbed corpus, we identify noun phrases containing multiple modifiers of interest (demonstratives, numerals, or adjective blocks) using Stanza dependency parses \citep{qi2020stanza}. These sentences are perturbed by decomposing each multi-modifier NP into variants containing only a single modifier per noun. Figure~\ref{pipeline} shows examples of how the perturbation process works for a typical sentence with two modifiers. Additional perturbation details are provided in Appendix~\ref{sec:perturbing}.

\subsection{Model Architecture}
We use decoder-only Transformer models based on the OPT model \citep{Zhang.etal2022}. We train three model sizes (52M, 110M, and 350M parameters) to examine whether model scale affects scope-homomorphic generalization. Architectural details are summarized in Table~\ref{tab:architecture}.

\begin{table}[t]
\centering
\resizebox{\columnwidth}{!}{%
\begin{tabular}{lccccc}
\toprule
\textbf{Model} & \textbf{Params} & \textbf{Layers} & \textbf{Hidden} & \textbf{Heads} & \textbf{FFN} \\
\midrule
Small  & 52M  & 8  & 768  & 8  & 768  \\
Medium & 110M & 12 & 768  & 12 & 3072 \\
Large  & 350M & 24 & 1024 & 16 & 4096 \\
\bottomrule
\end{tabular}
}
\caption{Architecture details for the OPT-based decoder-only Transformer models used in our experiments.}
\label{tab:architecture}
\end{table}

\subsection{Training}
Models are trained for 40 epochs with batch size 8 using AdamW \citep{Loshchilov.Hutter2017} and a learning rate of $5\times10^{-5}$. We select the checkpoint with the lowest validation loss and train five models per size using different random seeds. The full hyperparameter configuration is provided in Appendix \ref{sec:hyperparameters}.

\begin{figure*}[ht]
  \includegraphics[width=0.48\linewidth]{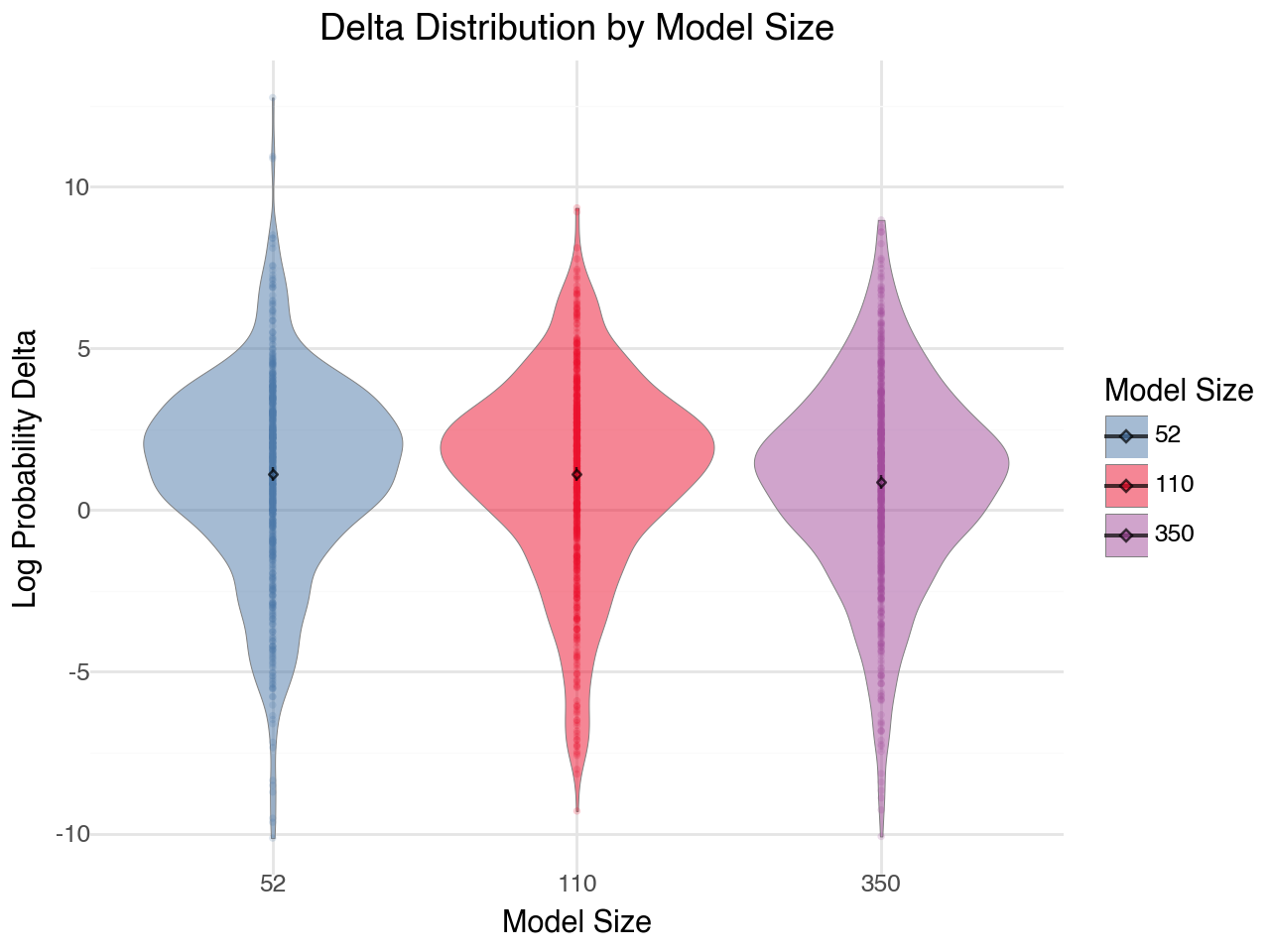} \hfill
  \includegraphics[width=0.48\linewidth]{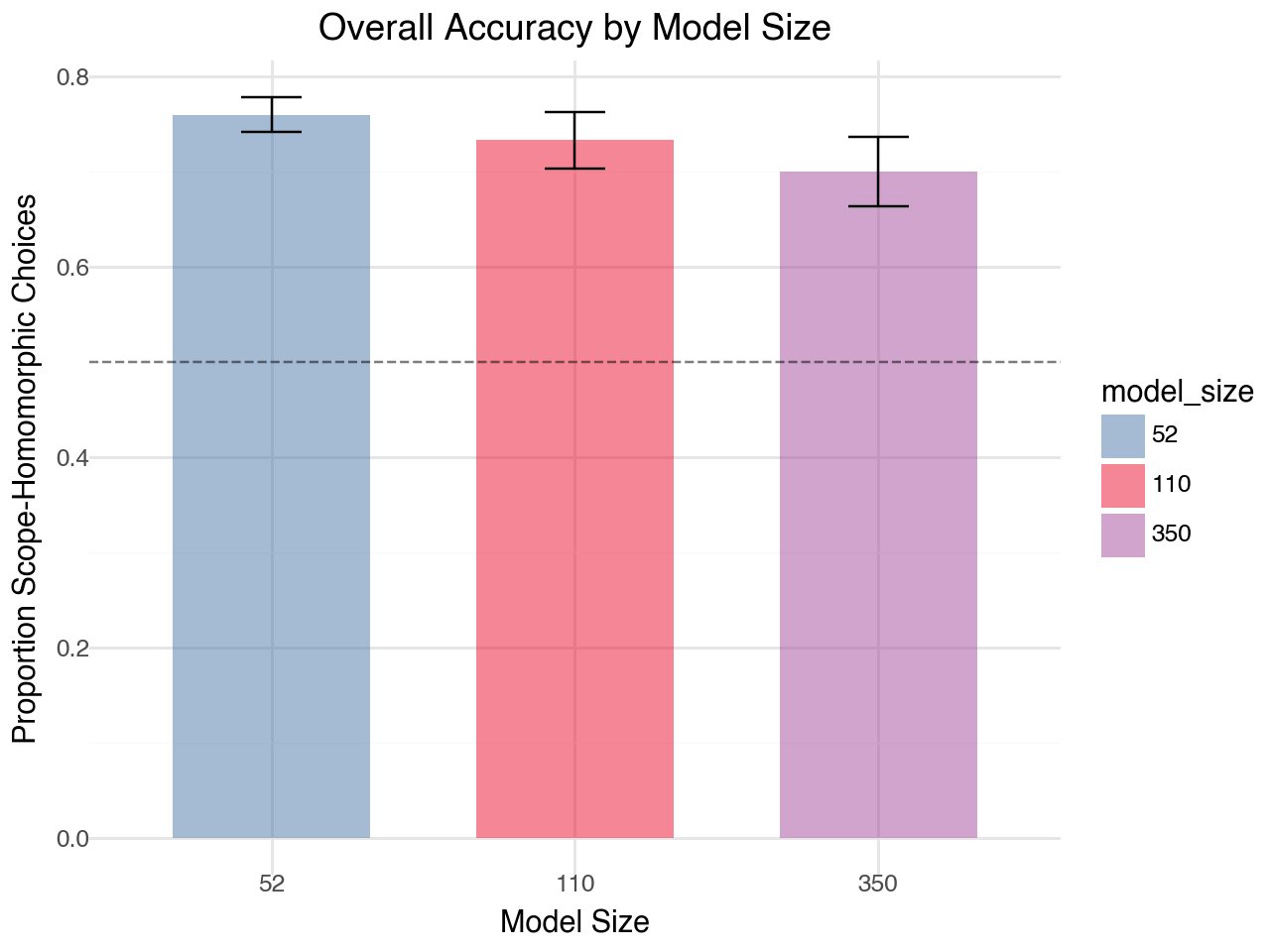}
  \caption {Overall contextual delta and accuracy for homomorphic word orders across model sizes. All three models exhibited positive mean $HC\Delta$ values and contextual accuracies above chance, indicating a reliable preference for homomorphic word orders. Error bars represent 95\% confidence intervals.}
  \label{fig:dataset_sum}
\end{figure*}

\subsection{Evaluation Corpus}
We evaluate models using minimal pairs following prior work on LM grammaticality evaluation \citep{Linzen.etal2016a,Marvin.Linzen2018}. Each candidate multi-modifier NP produces a pair consisting of the original and a version with swapped modifier order (see Figure~\ref{pipeline}). For each modifier combination (Dem-Num, Dem-Adj, Num-Adj), we sample 50 pairs (25 from training and 25 from test data), yielding 150 evaluation items. The evaluation pairs sourced from training data will still be unseen, since they have multiple modifiers, but each single modifier version will have been seen in training; for samples sourced from test, the individual modifiers may not have been seen. 

\subsection{Evaluation}

Building on prior work measuring LM word order preferences in noun phrases \citep{Misra.Mahowald2024, Jumelet.etal2026}, we quantify homomorphic order preference by comparing the log probability of the canonical order with that of a swapped order. Therefore, for any two modifiers, $M_{1}$ and $M_{2}$ and a noun head $N$, we compute the magnitude of the model's homomorphic consistency (HC) conditioned on the context (e.g. the previous tokens), $c$, as the following:
\begin{align*}
HC\Delta(M_{1}M_{2}N | c) =&  \log P(M_{1}M_{2}N | c) \\
& - \log P(M_{2}M_{1}N | c)
\end{align*}
We additionally report the proportion of items where the model prefers the homomorphic order, where $C$ is the evaluation corpus:
$$
HC\%(C) = \frac{|\{c \in C \mid HC\Delta(M_{1}M_{2}N | c) > 0\}|}{|C|} 
$$

\section{Results}
\label{sec:results}

We evaluate models using three metrics: perplexity, preference magnitude ($HC\Delta$), and preference accuracy ($HC\%$). We first compare models across sizes, then examine differences between modifier combinations.

\subsection{Perplexity}

Surprisingly, perplexity did not improve with model size. On the test corpus, mean perplexity increased slightly from 86.95 for the 52M model to 87.56 for the 110M model and 88.46 for the 350M model, with pairwise $t$-tests showing that all three model sizes differed significantly from one another (52M vs. 110M: $t = -19.55$, $p < .001$; 52M vs. 350M: $t = -75.26$, $p < .001$; 110M vs. 350M: $t = -63.84$, $p < .001$).

We also measured perplexity on the evaluation corpus separately for homomorphic and non-homomorphic sentences. All models assigned lower perplexity to homomorphic orders (52M: 141.02; 110M: 138.55; 350M: 143.71) than to non-homomorphic orders (52M: 155.17; 110M: 150.92; 350M: 152.90), providing initial evidence for an ordering preference across model sizes. 

\subsection{Across Model Sizes}

The homomorphic consistency delta ($HC\Delta$) measures the strength of the model's preference, with positive values indicating a preference for the homomorphic order, while $HC\%$ measures the proportion of items where the model selects the homomorphic order. Figure~\ref{fig:dataset_sum} summarizes these metrics across model sizes.

All models significantly preferred homomorphic orders, with mean $HC\Delta$ values of 1.12, 1.11, and 0.87 and accuracies of 76.0\%, 73.3\%, and 70.0\% for the 52M, 110M, and 350M models, respectively. T-test showed, for all model sizes, both metrics were significantly above chance. These results indicate that all model sizes learned a preference for scope-homomorphic ordering, with no significant variation across model sizes on either metric.

\begin{figure*}[ht]
  \includegraphics[width=0.48\linewidth]{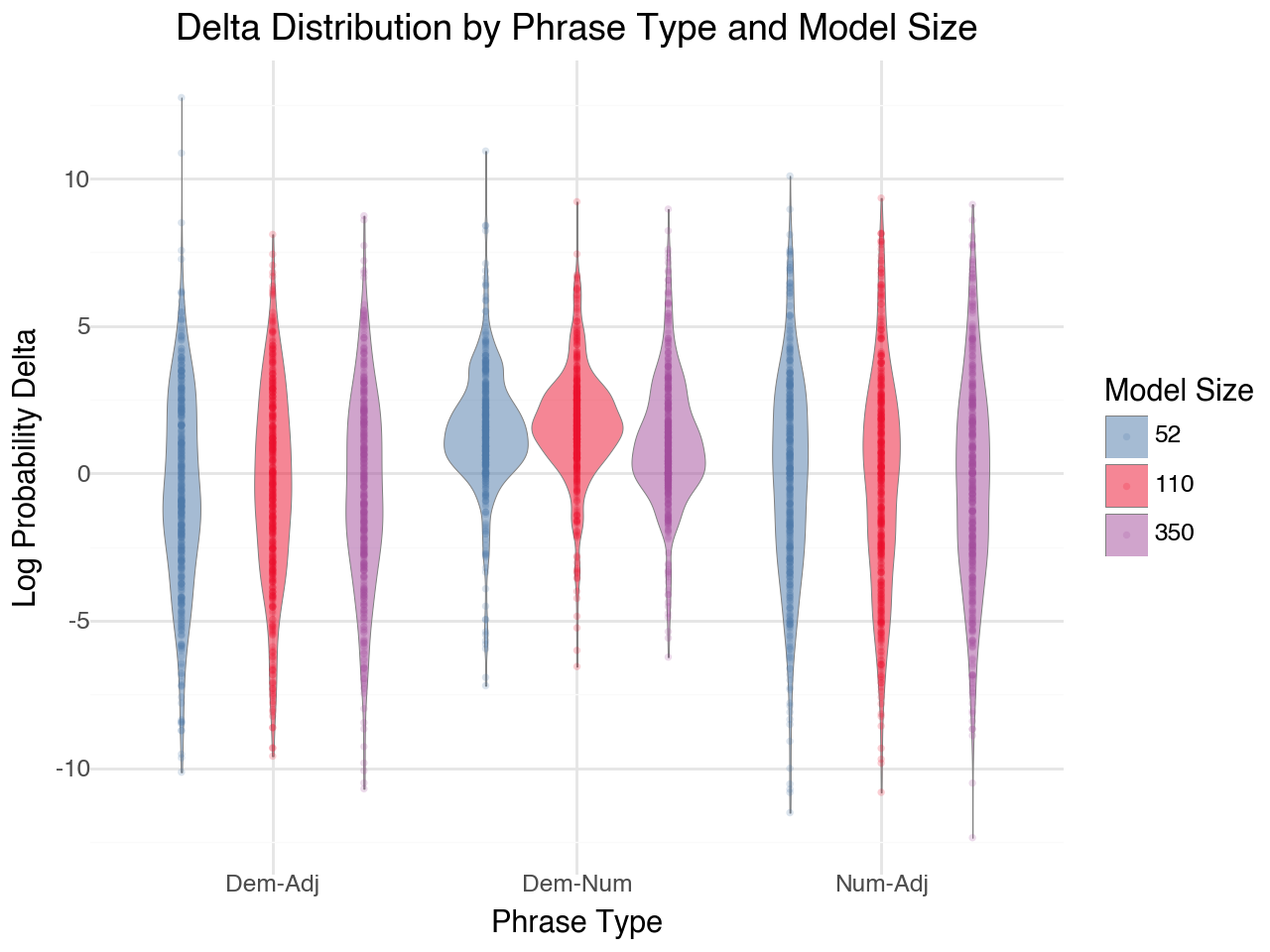} \hfill
  \includegraphics[width=0.48\linewidth]{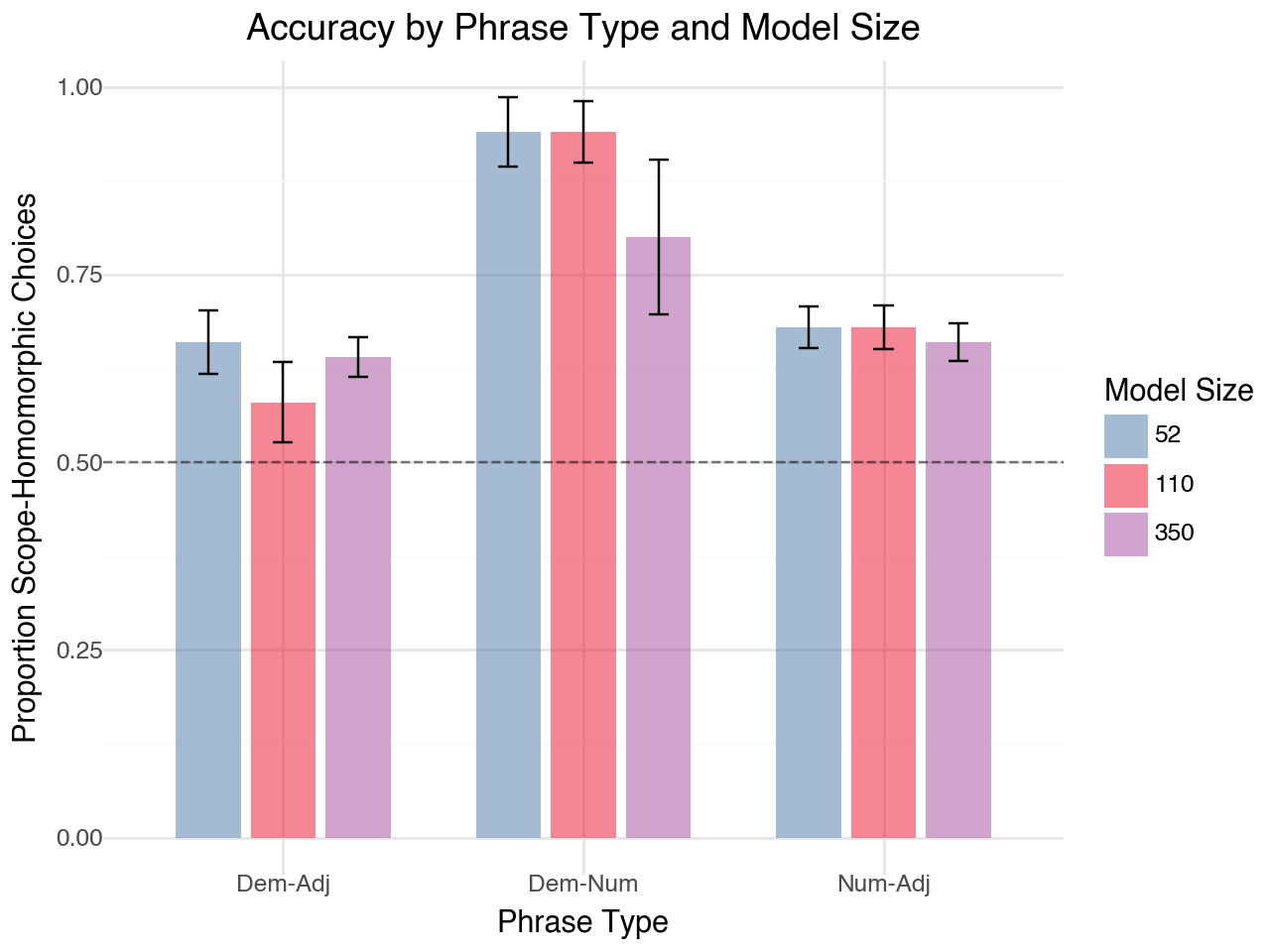}
  \caption {Homomorphic consistency elta and accuracy for homomorphic word orders by modifier type across model sizes. Dem-Num constructions elicited the strongest preference for homomorphic word orders across all three models, whereas Dem-Adj and Num-Adj constructions showed weaker but generally reliable preferences. Error bars represent 95\% confidence intervals.}
  \label{fig:by_type}
\end{figure*}

\subsection{By Modifier Combination}

\subsubsection{Summary Statistics}

A by-combination analysis revealed consistent differences across modifier constructions (Figure~\ref{fig:by_type}). Across all model sizes, Dem-Num constructions showed the strongest preference for homomorphic orders, with the highest $HC\Delta$ and accuracy values (52M: $HC\Delta$ = 2.07, $HC\%$ = 94\%; 110M: $HC\Delta$ = 2.12, $HC\%$ = 94\%; 350M: $HC\Delta$ = 1.51, $HC\%$ = 80\%). In contrast, Dem-Adj constructions showed the weakest preferences, while Num-Adj constructions exhibited intermediate effects. Most modifier combinations showed preferences significantly above chance, with the exception of the 52M model's Dem-Adj $HC\Delta$ effect. Overall, the relative ordering of modifier preferences was consistent across model sizes, although the 350M model showed weaker effects overall.

\subsubsection{Mixed Effects Model}
To examine whether preference strength varied systematically with model size and modifier type, we fit a mixed-effects regression predicting $HC\Delta$ from model size, modifier type, and their interaction, with random intercepts for model seed and evaluation item. Modifier type was treatment-coded with Dem-Num as the reference condition. Model size was treated as a continuous predictor to test for a linear trend across scales. Results are shown in Table~\ref{tab:mixed_effects_model}.

\begin{table}[ht]
\centering
\begin{tabular}{lrrrr}
\toprule
\textbf{Predictor} & \textbf{$\beta$} & \textbf{SE} & \textbf{$p$} \\
\midrule
Intercept & 2.255 & 0.184 & $< .001$ \\
Dem-Adj & -1.747 & 0.260 & $< .001$ \\
Num-Adj & -1.454 & 0.260 & $< .001$ \\
Size & -0.002 & $ 0.001$ & $< .001$ \\
Size $\times$ Dem-Adj & 0.002 & 0.001 & .002 \\
Size $\times$ Num-Adj & 0.002 & 0.001 & .003 \\
\bottomrule
\end{tabular}
\caption{Results of Linear mixed-effects model predicting preference strength ($HC\Delta$).}
\label{tab:mixed_effects_model}
\end{table}

The model revealed a significant effect of modifier type: both Dem-Adj and Num-Adj constructions showed weaker preferences than Dem-Num constructions. Model size had a very small negative effect on Dem-Num preferences, with significant but equally small interactions indicating that this relationship was specific to the Dem-Num condition. Overall, modifier type explained much more variation in preference strength than model size.

\subsection{Summary of Findings}

Across analyses, all three model sizes showed reliable homomorphic consistency, with both preference strength ($HC\Delta$) and accuracy ($HC\%$) significantly above chance. The magnitude of this effect did not significantly vary with model size. A consistent effect of modifier type was also observed, with Dem-Num constructions producing the strongest homomorphic preferences across models, while Dem-Adj and Num-Adj constructions showed weaker effects overall. This result is in contrast to human behaviour where Dem-Adj typically shows the strongest preference. These patterns were supported by the mixed-effects model, which indicated that differences across modifier types were systematic and that the reduction in effect with increasing model size was driven primarily by Dem-Num constructions. All effects of model size are quite small, suggesting negligible differences in preferences across model sizes.

\section{The Source of the Preference}
\label{sec:bias-source}

To better understand the source of the models' ordering preferences, we investigate whether their behaviour can be explained by distributional factors present in the training corpus. Specifically, we examine two potential sources of bias: the frequency of different modifier types in the training data and the strength of noun-modifier associations.

\subsection{Frequency-Based Sources}
We counted the number of noun phrases containing each single modifier type in order to determine whether any specific dependency type was substantially overrepresented in the training data. We find that adjectival dependencies were the most common type in the training corpus. Table \ref{tab:modifier_counts} lists the total counts. Overall, these findings suggest that the models’ performance on Dem-Num word order cannot be straightforwardly attributed to simple frequency-based biases in the training corpus. 

\begin{table}[h]
\centering
\begin{tabular}{lr}
\toprule
\textbf{Modifier Type} & \textbf{Count} \\
\midrule
Num & 1,266,334 \\
Adj & 4,185,588 \\
Dem & 124,234 \\
\bottomrule
\end{tabular}
\caption{Counts by modifier type in training corpus.}
\label{tab:modifier_counts}
\end{table}

We also searched for noun phrases in which a numeral served as the head and was modified by a demonstrative (e.g., \textit{Those two were very fast.}), as these would not have been removed by our filter and could have provided unintended evidence for Dem-Num ordering. We found no instances of this construction in the training corpus, ruling it out as a source of the observed bias.

We also examined postnominal modifiers, since constructions involving adjectives generally produced weaker preferences in our experiments. Although English strongly favours prenominal modifiers, the corpus contains 29,725 postnominal adjective constructions (e.g.`Billboard magazine reviewed the \textit{project favorable} in its October 1967 issue.') and 65,773 postnominal numerals (e.g. `step 5'). In contrast, there are no instances of postnominal demonstratives. This distributional asymmetry may contribute to the strong Dem-Num preference, but it cannot fully explain the observed modifier hierarchy, since it predicts stronger preferences for both Dem-Num and Dem-Adj relative to Num-Adj. See Table \ref{tab:postnominal_counts} for these counts. 

\begin{table}[h!]
\centering
\begin{tabular}{lc}
\toprule
\textbf{Modifier Type} & \textbf{Postnominal Instances} \\
\midrule
Dem    & 0 \\
Adj   & 29{,}725 \\
Num & 65{,}773 \\
\bottomrule
\end{tabular}
\caption{Post-nominal modifier counts in training data.}
\label{tab:postnominal_counts}
\end{table}

\subsection{Pointwise Mutual Information}
Previous work has proposed that human scope-homomorphic preferences may arise from tracking co-occurrence statistics: because adjectives are more strongly associated with nouns than numerals or demonstratives, learners may prefer to place them closer to the noun \citep{Culbertson.etal2020}. This association can be quantified using pointwise mutual information.

\subsubsection{Corpus Analysis}

To investigate whether the models' modifier ordering preferences could be explained by distributional properties of the training corpus, we computed PMI scores between nouns and their modifiers in our training corpus. PMI was computed separately for each dependency type (amod, nummod, det=demonstrative) using the following equation,  following \citet{Culbertson.etal2020}:
$$
\mathrm{PMI}(m, n, t)
=
\log
\frac{p(n,m \mid t(n,m))}
     {p(n \mid t(n,m))\,p(m \mid t(n,m))}
$$
where \textit{n} is a noun, \textit{m} is a modifier, and \textit{t(n,m)} is the dependency relation connecting them. Thus, we computed PMI scores over the training corpus, estimating modifier probabilities relative to other modifiers of the same dependency type and noun probabilities relative to nouns occurring with that modifier type. 

Table~\ref{tab:pmi_by_type} reports average PMI scores by dependency type. Consistent with previous work \citep{Culbertson.etal2020}, adjectives showed the highest noun-association strength, followed by numerals and demonstratives, suggesting that distributional association could serve as a basis for making generalizations. 

\begin{table}[t]
\centering
\begin{tabular}{lcc}
\toprule
\textbf{Modifier Type} & \textbf{Training} & \textbf{Evaluation} \\
\midrule
Adj (\textsc{amod})     & 3.24 & 3.05 \\
Num (\textsc{nummod})     & 2.59 & 1.58 \\
Dem (\textsc{det})  & 0.82 & 1.07 \\
\bottomrule
\end{tabular}
\caption{Average PMI scores by modifier type in the training corpus and evaluation items. Adjectival modifiers exhibit the highest average PMI values, while demonstratives exhibit the lowest in both the training corpus and evaluation sentences matching the suggested effect of PMI on scope-homomorphism.}
\label{tab:pmi_by_type}
\end{table}

\subsubsection{Predicting Model Behaviour}

To test whether these associations predicted model behaviour, we conducted an item-level analysis relating PMI differences to LM ordering preferences. For each evaluation item, we computed the difference between the PMI of the noun-adjacent modifier and the more distant modifier.

Thus, for a homomorphic order $M_1 M_2 N$, we calculate:
$$
    \Delta_{\mathrm{PMI}}(M_1 M_2 N) = \mathrm{PMI}(M_2,N) - \mathrm{PMI}(M_1,N)
$$
A positive $\Delta_{\mathrm{PMI}}$ indicates that the modifier closer to the noun has a stronger noun association, predicting a preference for the homomorphic order. We compare this prediction with LM preferences measured by $HC\Delta$, evaluating both their correlation and their agreement across items.

Across all three model sizes, PMI differences showed weak negative relationships with model preferences. Correlations were small for the 52M and 110M models (Pearson's $r = -0.094$ and $-0.105$, respectively; both $p < .05$) and non-significant for the 350M model ($r = -0.062$, $p = .114$). Agreement between PMI-based predictions and model preferences was near chance across all models (52M: 53.1\%; 110M: 52.2\%; 350M: 54.4\%), suggesting that noun–modifier association strength did not explain the observed ordering preferences.

If models' preferences were primarily driven by lexical co-occurrence statistics, we would expect stronger agreement between PMI-based predictions and model preferences. Instead, PMI predictions were only marginally above chance and showed a weak negative relationship with model preferences. Thus, while corpus-level PMI reflects modifier ordering tendencies, modifier–noun association strength alone cannot explain the generalizations learned by the models. Overall, models' ordering behaviour seems to be shaped by additional syntactic, semantic, or structural generalizations that are not captured by PMI.

\section{Discussion and Future Work}
Our results demonstrate that language models trained under artificial language learning conditions develop a reliable preference for scope-homomorphic noun phrase orders despite never observing noun phrases containing multiple modifiers during training. Across all model sizes, models consistently preferred homomorphic over non-homomorphic alternatives, showing that this generalization can emerge from impoverished input without explicitly encoded linguistic knowledge. In this respect, our findings parallel the central result of human ALL studies, where learners similarly extrapolate beyond the evidence provided during training \citep{Culbertson.Adger2014,Martin.etal2020}. While this does not demonstrate that humans acquire the bias through the same mechanism, it shows that strong innate constraints are not necessary for this preference to emerge and motivates further investigation into the statistical and structural properties of the input that support such learning.

Our findings raise new questions about the representations underlying these generalizations. If lexical association statistics are insufficient, as suggested by our PMI analysis, models may instead be relying on more abstract structural information acquired during training. Future work could investigate this possibility by training linear probes on frozen model representations to test whether syntactic head-modifier relationships learned from single-modifier noun phrases can be recovered in unseen multi-modifier structures. If such representations are present, their relationship with model ordering preferences could reveal whether structural knowledge, rather than lexical association alone, supports scope-homomorphic generalization.

A second direction for future work concerns the role of distributional statistics in human language learning. Previous work has shown that corpus-derived PMI reflects the scope-homomorphic ordering preference observed across languages \citep{Culbertson.etal2020}. However, it remains unclear whether humans actually rely on these distributional associations when learning or generalizing word order. Our results suggest that the language models do not. This raises the question of whether PMI is genuinely operative in human language learning or merely correlates with the observed ordering preferences. Developing methods to measure or manipulate the influence of such statistics in human ALL experiments would therefore be an interesting direction for future work.

Taken together, these findings demonstrate the value of using controlled language model training as a tool for investigating the origins of linguistic biases. By placing models under learning conditions that approximate the underdetermined input available in human artificial language learning experiments, we can begin to separate what aspects of linguistic generalization require language-specific constraints from what can emerge through experience. While our results do not resolve the specific source of scope-homomorphic preferences, they show that such biases can arise from impoverished input and provide a framework for further investigating the representations and learning mechanisms that support them.

\section*{Limitations}

Here, we note a few methodological limitations of our experiments. First, while language models provide a useful framework for investigating whether linguistic generalizations can emerge from impoverished input, they are not direct models of human language learners. LMs differ from humans in their learning objectives, input distribution, processing constraints, and representational capacities. Therefore, the emergence of scope-homomorphic preferences in LMs does not demonstrate that humans acquire this bias through the same mechanisms. Rather, our results demonstrate that such preferences can emerge in a general learning system without direct evidence for the relevant structure, providing a test of what types of biases are possible under controlled learning conditions.

Second, although our training environment removes direct evidence for multi-modifier noun phrase ordering, it is not equivalent to the ALL conditions used in human experiments. In human ALL studies, participants are exposed to a small artificial language that is designed to isolate a specific learning problem, but their behaviour reflects inductive biases acquired through prior experience with their native language or some innate bias. On the other hand, our models are trained on a broader linguistic environment that may contain indirect evidence about modifier relationships. This difference is not a limitation for our primary goal: we do not aim to replicate the process by which humans acquire scope-homomorphic preferences, but rather to test whether such preferences can be recovered when direct evidence for the relevant structure is unavailable. Thus, while our results cannot be interpreted as a direct model of the mechanisms underlying human behaviour during an ALL study, they provide a comparable test of whether scope-homomorphic generalizations can emerge in the absence of direct evidence for the relevant structure.

Finally, our investigation is limited to English, despite scope-homomorphism being proposed as a cross-linguistic generalization. For example, in a sample of 576 languages, 472 were found to follow scope-homomorphic modifier ordering patterns, corresponding to approximately 82\% of the languages surveyed \citep{Dryer2018}. While English provides a useful test case due to the availability of corpora and well-studied modifier ordering patterns, future work should examine whether similar preferences emerge in models trained on other languages with different modifier distributions.



\bibliography{custom}
\newpage
\appendix

\section{Perturbing the Corpus}
\label{sec:perturbing}

To construct the perturbed training corpus, we first parsed each sentence using a dependency-based representation and extracted noun phrases together with their associated modifiers. For each noun, we identified three types of nominal modifiers: demonstrative determiners (e.g., this, that), numerals (via nummod relations), and adjectival modifiers (via amod relations). Adjectival modifiers were combined with any adverbial modification (via advmod) of the adjective to preserve multi-word adjectival phrases (e.g., very large). Following previous work, non-demonstrative determiners are excluded from this modifier count. Adjective sequences and compound noun phrases are treated as single blocks.

Once modifiers were extracted, we constructed alternative versions of each sentence by recombining only one of the available modifiers for a given noun phrase while preserving the rest of the sentence structure. This allowed us to generate multiple surface variants of a single input sentence corresponding to different modifiers from the same underlying modifier set.

One issue with Stanza's dependency parses is that it systematically misparses clause initial "that one", often treating "that" as a subordinating conjunction (SCONJ) rather than part of a demonstrative nominal construction.To avoid propagating these parsing errors into the rest of the process, we explicitly filtered out sentences containing this pattern by detecting occurrences where "that" is tagged as a SCONJ and is immediately followed by one.

In total, 6,032,739 sentences were processed. Of these, 148,367 contained complex noun phrases and were filtered out, leaving 5,884,372 sentences with simple noun phrase structure to be included in the perturbed corpus. Once filtered out, the complex NP sentences were perturbed which created a set of 664,272 rewritten sentence variants for the corpus.

\section{Training Hyperparameters}
\label{sec:hyperparameters}

\begin{table}[h]
\centering
\begin{tabular}{ll}
\hline
\textbf{Hyperparameter} & \textbf{Value} \\
\hline
Optimizer & AdamW  \\
Learning rate & $5 \times 10^{-5}$ \\
Adam $\beta_1$ & 0.9 \\
Adam $\beta_2$ & 0.999 \\
Adam $\epsilon$ & $1 \times 10^{-8}$ \\
Learning rate scheduler & Linear \\
Number of training epochs & 40 \\
Per-device training batch size & 8 \\
Per-device evaluation batch size & 8 \\
Gradient accumulation steps & 1 \\
Dataloader num workers & 8 \\
Evaluation strategy & Steps \\
Evaluation interval & 500 steps \\
Save strategy & Steps \\
Mixed precision (fp16) & True \\
Ignore data skip & False \\
Seed & 37,42,78,414,755 \\
\hline
\end{tabular}
\caption{Training hyperparameters used for all models.}
\label{tab:training_hyperparams}
\end{table}

\end{document}